\documentclass[sigconf, nonacm]{acmart}

\AtBeginDocument{%
  }

\usepackage{minted}
\setminted{
    breaklines=true,
    breaksymbolleft= 
}
\usepackage{algorithm}
\usepackage{algpseudocode}
\usepackage{amsmath}
\usepackage{cleveref}
\usepackage{tikz}
\usetikzlibrary{arrows.meta, positioning, fit, backgrounds, calc}
\usepackage{placeins}

\usepackage[utf8]{inputenc}
\usepackage{newunicodechar}
\newunicodechar{μ}{\textmu}
\newunicodechar{Σ}{$\Sigma$}
\newunicodechar{ᵢ}{$_i$}
\newunicodechar{ᶜ}{$^c$}

\title{FlashVector: Agent for Hierarchical Model Serving Stack Optimization}

\author{%
  \text{Qi Wu}$^{1,2}\quad$ 
  \text{Lohan Lemire}$^{2}\quad$ 
  \text{Kai Meng}$^{2}\quad$ 
  \text{Zhongmou Cai}$^{2}\quad$   
  \text{Raphael Bargues}$^{2}\quad$     
  \text{Petr Zhitnikov}$^{2}\quad$      
  \text{Zeyuan Cao}$^{2}\quad$     
  \text{Yao Wang}$^{2}\quad$   
  \text{Shujun Bian}$^{2}\quad$
  \text{Wei Chen}$^{2}\quad$  
  \text{Sean Sheng}$^{2}\quad$    
  \\ \vspace{2pt}
  \small $^{1}$Stanford University
  \small $^{2}$Unity Vector AI Team
}

\date{August 2026}

\begin{document}

\fancyhead{}

\begin{abstract}
Model serving is one of the largest cost drivers in production recommender systems. Maximizing its throughput requires navigating a deeply layered hierarchy: GPU kernels, the ML-framework computation graph, the model server, and on-demand feature processing—each demanding specialized domain expertise. Such cross-layer expertise is inherently difficult to acquire, and does not scale with a workload that continuously grows and evolves, leaving significant cost efficiency gains unrealized. While recent AI agents have demonstrated human expert level efficiency in standalone GPU kernel optimization, automated tuning and optimization for the rest of the serving stack remain largely unexplored. We present FlashVector, an agentic system that optimizes performance across all layers of the model serving stack. The key contribution is an extensible framework to generalize the single kernel optimization agent paradigm to heterogeneous technical stacks, and to deliver performance improvements holistically. After deployment in Unity's Vector advertising platform, FlashVector achieved up to \textbf{2$\times$} throughput increase and up to \textbf{1.98$\times$} latency speedup on model server, and up to \textbf{1.6$\times$} throughput increase on feature store. These optimizations were discovered not only at the GPU kernel and computation graph levels, but also across the other components of the model serving stack, such as the model server (NVIDIA Triton's C++ codebase) and the on-demand feature transformation service (Python codebase), demonstrating the extensibility of the framework to more complex system architectures.

\end{abstract}

\maketitle

\section{Introduction}\label{sec:intro}
Deep learning models are an essential component of modern recommender systems \cite{he2014practical} \cite{mcmahan2013ad}. At Unity's Vector advertising platform, DNN models are used throughout the funnel to retrieve candidates, estimate probability of conversion, and predict the long term monetization value. As one of the leading mobile game advertising solutions, Vector operates at scale. It manages hundreds of DNN models, and serves hundreds of thousands of inference requests per second in production.

\begin{figure}[!htbp]
    \centering
    \resizebox{\linewidth}{!}{%

\begin{tikzpicture}[
  font=\sffamily\scriptsize,
  every node/.style={align=center},
  box/.style={rectangle, rounded corners=4pt, draw, line width=0.8pt,
    align=center, inner xsep=7pt, inner ysep=6pt},
  flow/.style={->, line width=1.0pt},
]

\definecolor{irText}{HTML}{1F2933}
\definecolor{irNeutralFill}{HTML}{F7F9FC}
\definecolor{irNeutralBorder}{HTML}{CBD5E1}
\definecolor{irKnowledgeFill}{HTML}{F3F0FF}
\definecolor{irKnowledgeBorder}{HTML}{8B7BD6}
\definecolor{irCandidateFill}{HTML}{FFF4E8}
\definecolor{irCandidateBorder}{HTML}{D9A05B}
\definecolor{irValidationFill}{HTML}{EEF2FF}
\definecolor{irValidationBorder}{HTML}{7687D9}
\color{irText}

\node[box, fill=irValidationFill, draw=irValidationBorder,
  minimum width=4.6cm, minimum height=3.4cm] (modelserver) at (0,0) {};
\node[anchor=north west, font=\sffamily\bfseries\small]
  at ($(modelserver.north west)+(0.15,-0.15)$) {Model Server};

\node[box, fill=irKnowledgeFill, draw=irKnowledgeBorder,
  minimum width=3.6cm, minimum height=2.15cm]
  (mlframework) at ($(modelserver.center)+(0,-0.35)$) {};
\node[anchor=north west, font=\sffamily\bfseries\scriptsize]
  at ($(mlframework.north west)+(0.12,-0.12)$) {ML Framework};

\node[box, fill=irCandidateFill, draw=irCandidateBorder,
  minimum width=2.5cm, minimum height=0.95cm]
  (gpukernels) at ($(mlframework.center)+(0,-0.3)$) {};
\node[anchor=north west, font=\sffamily\bfseries\scriptsize]
  at ($(gpukernels.north west)+(0.1,-0.1)$) {GPU Kernels};

\node[box, fill=irNeutralFill, draw=irNeutralBorder,
  minimum width=2.6cm, minimum height=1.3cm, font=\sffamily\bfseries\small,
  left=2.6cm of modelserver] (adserver) {Ad Server};

\node[box, fill=irNeutralFill, draw=irNeutralBorder,
  minimum width=2.9cm, minimum height=1.8cm, anchor=north east]
  (featurestore) at ($(modelserver.north east)+(0,2.35)$) {};
\node[anchor=north west, font=\sffamily\bfseries\small]
  at ($(featurestore.north west)+(0.12,-0.12)$) {Feature Store};

\node[box, fill=irKnowledgeFill, draw=irKnowledgeBorder,
  minimum width=2.2cm, minimum height=0.7cm]
  (ondemand) at ($(featurestore.center)+(0,-0.3)$) {\scriptsize\bfseries Ondemand Feature};

\draw[flow, color=irNeutralBorder!80!black]
  (adserver.east) -- (modelserver.west);
\draw[flow, color=irNeutralBorder!80!black]
  (adserver.north) |- (featurestore.west);

\end{tikzpicture}
    }
    \caption{A typical model serving stack.}
    \label{fig:model_serving_example_stack}
\end{figure}
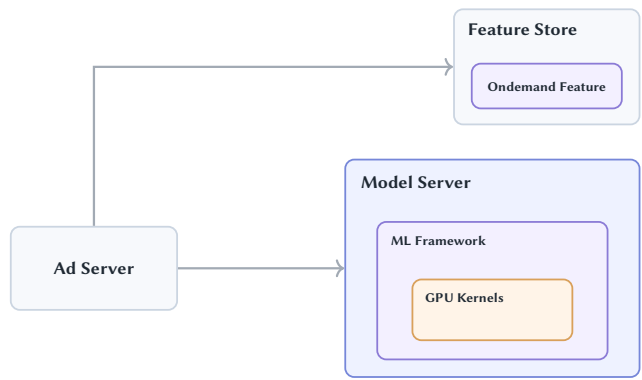

Serving models at this scale (tens of thousands of GPU and CPU nodes) requires a complex distributed infrastructure, with data crossing multiple network hops, nodes, and devices. In a typical model serving architecture (e.g.,  \Cref{fig:model_serving_example_stack}), clients split incoming requests into parallel calls to feature stores and model servers. Upon receiving feature-enriched requests, the model server employs dynamic batching strategies and parallel inference to maximize throughput. Once a batched request is generated, it is passed to an ML framework runtime (e.g., PyTorch), which traverses the execution graph and launches many GPU kernels that  perform the mathematical operations. Multiple programming languages are typically involved—Golang, C++, Python, and CUDA, among others. 

Optimizing this hierarchical serving stack has historically been challenging, as each layer demands distinct domain expertise. Client-side optimizations focus on request fan-out and parallelism. Within the model server, performance hinges on managing throughput-latency trade-offs through techniques such as dynamic batching and multi-process parallel inference, which interact directly with the ML framework runtime. Within the ML framework runtime, performance depends on proper computation graph configurations (e.g., PyTorch module structure, tensor layouts, operator selection, and minimizing host-device synchronization). Finally, low-level efficiency requires custom GPU kernel optimizations, such as kernel fusion, memory bandwidth tuning, and Tensor Core utilization. Without deep domain expertise across every layer, substantial cost efficiency gains often go unrealized.

Recent progress offers a partial answer to this gap. The emergence of kernel design agents \cite{ouyang2025kernelbenchllmswriteefficient} has significantly improved auto-generated GPU kernel efficiency, enabling teams without deep kernel expertise to deploy custom CUDA code. This motivates a broader question: Can we extend LLM-driven optimization from standalone GPU kernels to the entire model serving stack? In our experience, non-kernel layers contain just as many—if not more—optimization opportunities. A more efficient implementation within the model server or a better tuned runtime parameter can tremendously increase serving throughput. However, holistic stack optimization introduces new engineering challenges: discovering efficient code implementations, tuning the runtime parameter spaces, and finding optimal co-designs across both code and parameters.

In this paper, we present FlashVector, an agentic system that holistically optimizes the hierarchical model serving stack. This paper makes three contributions:

\begin{itemize}
\item \textbf{A layer agent abstraction.} Each serving layer plugs into the same optimization loop by conforming to a standard optimizing-agent interface: \textbf{Profile} (domain-specific tooling — Nsight for GPU kernels, eBPF for the model server), \textbf{Diagnose} (bottlenecks against a layer-specific knowledge base), \textbf{Optimize} (generate and apply the corresponding code or configuration change), and \textbf{Verify} (check the result against production evidence), plus a post-run \textbf{Refine} stage that writes findings back to the knowledge base.

\item \textbf{Optimize locally, verify globally.} FlashVector proposes optimizations independently within each layer of the stack — GPU kernels, the computation graph, the model server, feature processing — so a bottleneck in any one of them can be found and fixed on its own. But a candidate is accepted only when it is shown to improve the system as a whole: every change is measured against replayed production traffic and load held at the production latency SLO, and kept only if the end-to-end gain exceeds the measurement noise.

\item \textbf{An always-on optimization loop.} Serving optimizations decay with every retrain, traffic shift, and hardware refresh. Each run starts from the model's recorded history of accepted and rejected changes and writes its own outcome back, and the loop re-triggers itself automatically rather than waiting to be invoked.
\end{itemize}

The remainder of this paper is organized as follows. \Cref{sec:related_work} discusses related work. \Cref{sec:challengs} outlines the key challenges in extending kernel optimization agents to full serving stacks. \Cref{sec:core_system} details our system design and solution. \Cref{sec:case_studies} highlights several optimizations realized using FlashVector. Finally, \Cref{sec:discussion} discusses broader implications and concludes.

\section{Related Work}
\label{sec:related_work}

\textbf{Model Serving Systems.} A substantial body of systems work targets the model-serving layer itself, treating request routing, batching, and resource allocation as the object of optimization rather than the model's code. Clipper~\cite{crankshaw2017clipper} introduced a general-purpose low-latency prediction-serving layer with adaptive batching; TensorFlow Serving~\cite{olston2017tensorflow} established production-grade serving runtimes still widely deployed today. Subsequent systems targeted specific bottlenecks: Nexus~\cite{shen2019nexus} schedules many models across a GPU cluster for video analytics, Clockwork~\cite{gujarati2020serving} pursues predictable per-request latency through tight execution control, and InferLine~\cite{crankshaw2020inferline} provisions and auto-scales multi-stage prediction pipelines end to end. More recently, serving systems for large generative models—Orca~\cite{yu2022orca}, vLLM~\cite{kwon2025vllm}, and AlpaServe~\cite{li2023alpaserve}—introduced iteration-level scheduling, paged KV-cache memory management, and statistical multiplexing of model-parallel replicas, respectively. Closest to the on-demand feature-processing bottleneck in \Cref{sec:case_studies}, Willump~\cite{kraft2020willump} optimizes end-to-end ML inference pipelines whose bottleneck lies in feature computation rather than model execution, using statistically-motivated approximations. These systems share FlashVector's premise that serving-layer inefficiency is a first-class optimization target, but each hand-designs a fixed policy or architecture for one bottleneck class; FlashVector instead uses an LLM agent to discover analogous fixes automatically across an arbitrary and evolving stack, without a new system having to be designed for each new bottleneck.

\textbf{Serving-time Autotuning.} A parallel line of work automatically searches serving-time configuration spaces, closest to the parameter-tuning problem in \Cref{sec:case4}. MArk~\cite{zhang2019mark} and Cocktail~\cite{gunasekaran2022cocktail} jointly select hardware tier, model variant, and autoscaling policy to meet cost and SLO targets; Morphling~\cite{wang2021morphling} uses meta-learning over historical configurations to converge on near-optimal resource and batch settings for a new model with few trials; GSLICE~\cite{dhakal2020gslice} and DVABatch~\cite{cui2022dvabatch} search GPU spatial-partitioning and batch-composition policies, respectively, to raise multi-tenant throughput at fixed latency. Like our Case~4, these systems treat throughput-at-SLO as a black-box objective over a configuration space, but each is a purpose-built search procedure over a hand-scoped parameter set along one axis of the stack (hardware tier, batching, or GPU sharing). FlashVector instead lets a single agent's profile--diagnose--optimize--verify loop range over the same class of levers alongside code-level changes, and, as \Cref{sec:optimization_loop} describes, schedules itself continuously.

\textbf{Kernel Design Agent.} Much work (\cite{liao2026kernelevolvescalingagentickernel}, \cite{KernelAgent2025}, \cite{nagaitsev2025pike}, \cite{dong2026kernelblaster}, \cite{zhang2026accelopt}, \cite{hong2025autocomp}, \cite{dai2026cuda}, \cite{guo2025evoengineer}, \cite{li2025autotriton}, \cite{ding2026prompts}, \cite{kernel_design_agents_2026}, \cite{sglangteam2026agent}, \cite{liu2026drkernelreinforcementlearning}) have been exploring GPU kernel optimization as a task that large language model (LLM) agents can carry out largely autonomously. KernelBench~\cite{ouyang2025kernelbenchllmswriteefficient} first proposed a standard suite of PyTorch programs to assess coding LLMs, and showed frontier models still fall short without more sophisticated harness. KernelAgent~\cite{KernelAgent2025} and KDA~\cite{kernel_design_agents_2026} are multi-agent systems that demonstrated the profile--diagnose--optimize--verify loop and showed superior performance than standard compilation. FlashVector is a direct extension of the profiling guided paradigm: it reuses the same profile--diagnose--optimize--verify loop, but applies it to layers where standalone kernel-agent techniques do not transfer as-is — a model server's C++ request path, a feature-processing service's Python runtime, and pure runtime configuration — each requiring its own profiling tool, diagnosis context, and correctness bar rather than a single kernel's fixed input/output contract.

\section{Challenges}
\label{sec:challengs}

Extending automated optimization beyond standalone kernels to the full serving stack introduces four fundamental challenges. 

To begin with, the fragmented tooling and heterogeneity across the stack mask true performance bottlenecks. Each layer of the serving stack operates in a distinct ecosystem with its own implementation languages (e.g., Go/C++ in model servers vs. Python in framework runtime vs. CUDA/Triton in kernels) and specialized profiling tools (e.g., eBPF, PyTorch Kineto, Nsight Systems). Because performance signals and "optimization dialects" are completely fragmented across boundaries, few engineers possess the cross-layer domain expertise needed to localize the true bottleneck, let alone coordinate a fix across it.

Second, non-kernel layers present a joint optimization problem. Maximizing throughput requires simultaneously discovering efficient code (e.g., altering server-side serialization or graph structures) and tuning multi-dimensional parameter spaces (e.g., dynamic batching thresholds, thread pool allocations, and GPU queue depths). Navigating this combined, high-dimensional search space requires co-optimizing discrete code changes alongside continuous parameters. 

Furthermore, standalone kernel generation tasks operate on localized, single-file code snippets with predictable inputs and outputs. In contrast, full stack optimization requires reasoning over a multi-repository codebase—spanning distributed microservices, feature store, and model server. These multi-file edits cannot be evaluated in isolation: unlike offline micro benchmarks, candidate configurations and code changes across the stack must strictly preserve production service-level agreements (SLAs), such as tight $p_{99}$ tail-latency budgets, without triggering cascading downstream failures.

Performance optimization is also complicated by the fact that it is an inherently continuous, moving target rather than a static, one-off effort. Manual tuning relies on an iterative, slow cycle: capturing distributed traces, forming hypotheses, implementing code or parameter changes, redeploying, and re-benchmarking under load. In production platforms like Unity Vector, these manual interventions quickly decay. Models are continuously retrained and re-exported, feature pipelines evolve, traffic patterns shift diurnally, and the underlying fleet infrastructure (GPU generations, driver versions, and framework runtimes) is updated regularly. A manual optimization validated on one model release or traffic profile frequently becomes suboptimal—or even detrimental—under the next. Because human engineering teams cannot indefinitely sustain the high operational tax of manually re-evaluating the entire 4-layer stack, optimizations expire, leaving significant efficiency gains unrealized over time.

\begin{figure}[t]
  \centering
  \includegraphics[width=\linewidth]{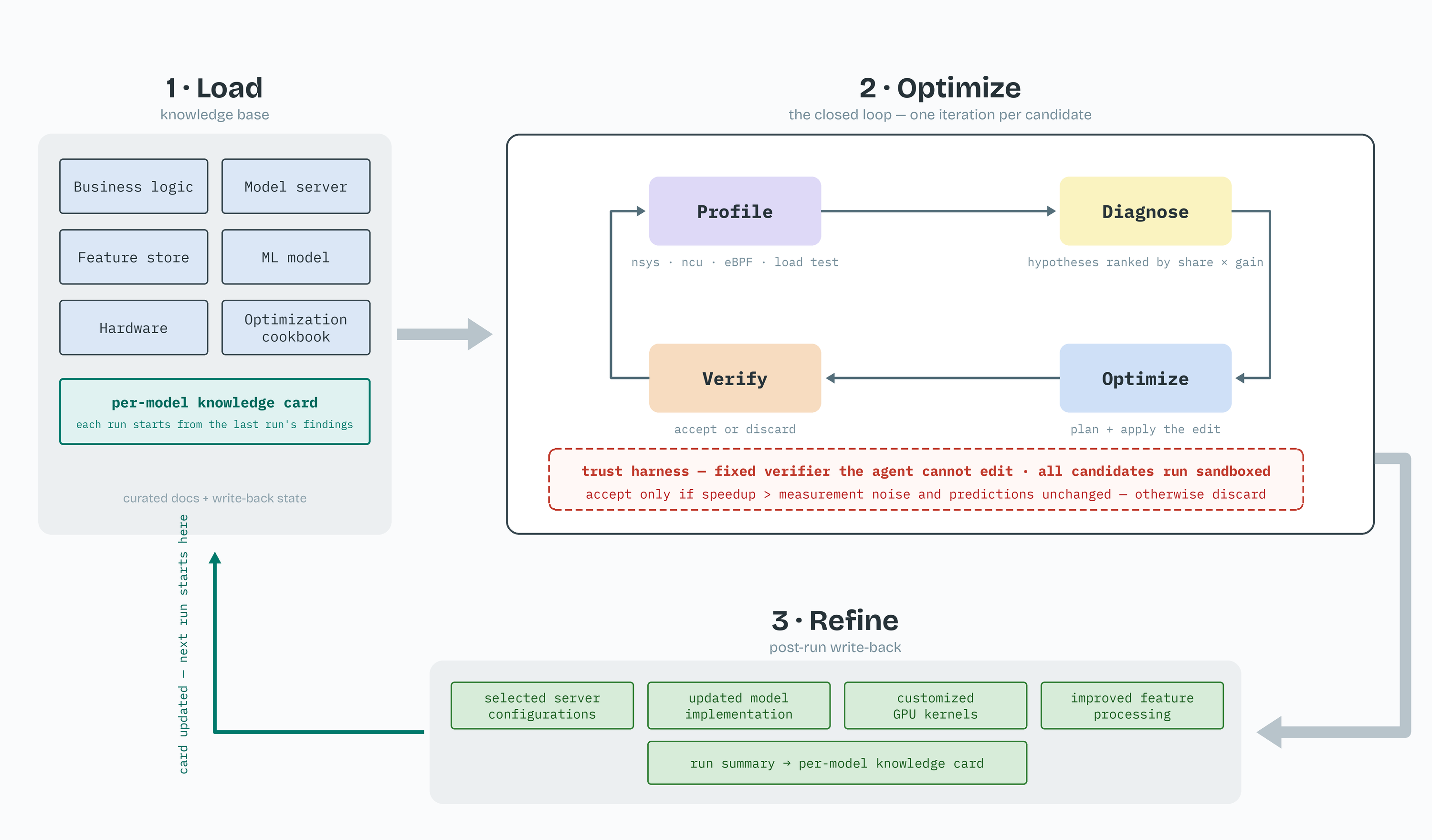}
  \caption{The three-stage workflow of FlashVector. }
  \label{fig:figure2}
\end{figure}

\section{Core System}
\label{sec:core_system}
A vanilla approach would feed an LLM official NVIDIA and PyTorch documentation, the source code of the models and model servers, and a high-level profiling report such as \texttt{torch.profiler}, then prompt it to produce optimizations in one shot. This does not work well, for two reasons.

First, generic documentation and high-level reports are necessary but insufficient context: as \Cref{sec:challengs} notes, each layer's performance signals are fragmented across specialized tools and ecosystems, so feeding the agent generic material either wastes its context window on irrelevant detail or omits the domain-specific signal it actually needs. In practice this means precise kernel traces (nsys/ncu) instead of \texttt{torch.profiler}'s coarser summaries, exact tensor shapes and layouts to curb hallucination during roofline analysis, and business-logic context — e.g., whether a feature is request- or candidate-level — that lets the agent propose optimizations that are difficult for traditional machine learning compiler to realize, such as model-level loop-invariant code motion.

Second, even with the right context, an LLM rarely reaches the correct optimization in a single attempt: it can misattribute a bottleneck from its own profile. Reliable progress requires an iterative loop with a harness that checks each step, rather than a one-shot generation — which is what the rest of this section describes.
  
At a high level, FlashVector has three stages, shown in \Cref{fig:figure2}: Knowledge base loading, Iterative optimization, and Knowledge refinement. The agent first retrieves domain-specific knowledge from pre-compiled reference documents for the targeted serving layer. Then it creates a closed feedback loop to iteratively profile, diagnose, optimize, and verify candidate setups. Finally, it updates the reference documents with the optimized full-stack configuration.

\begin{figure*}[t]
  \centering
  \includegraphics[width=\textwidth]{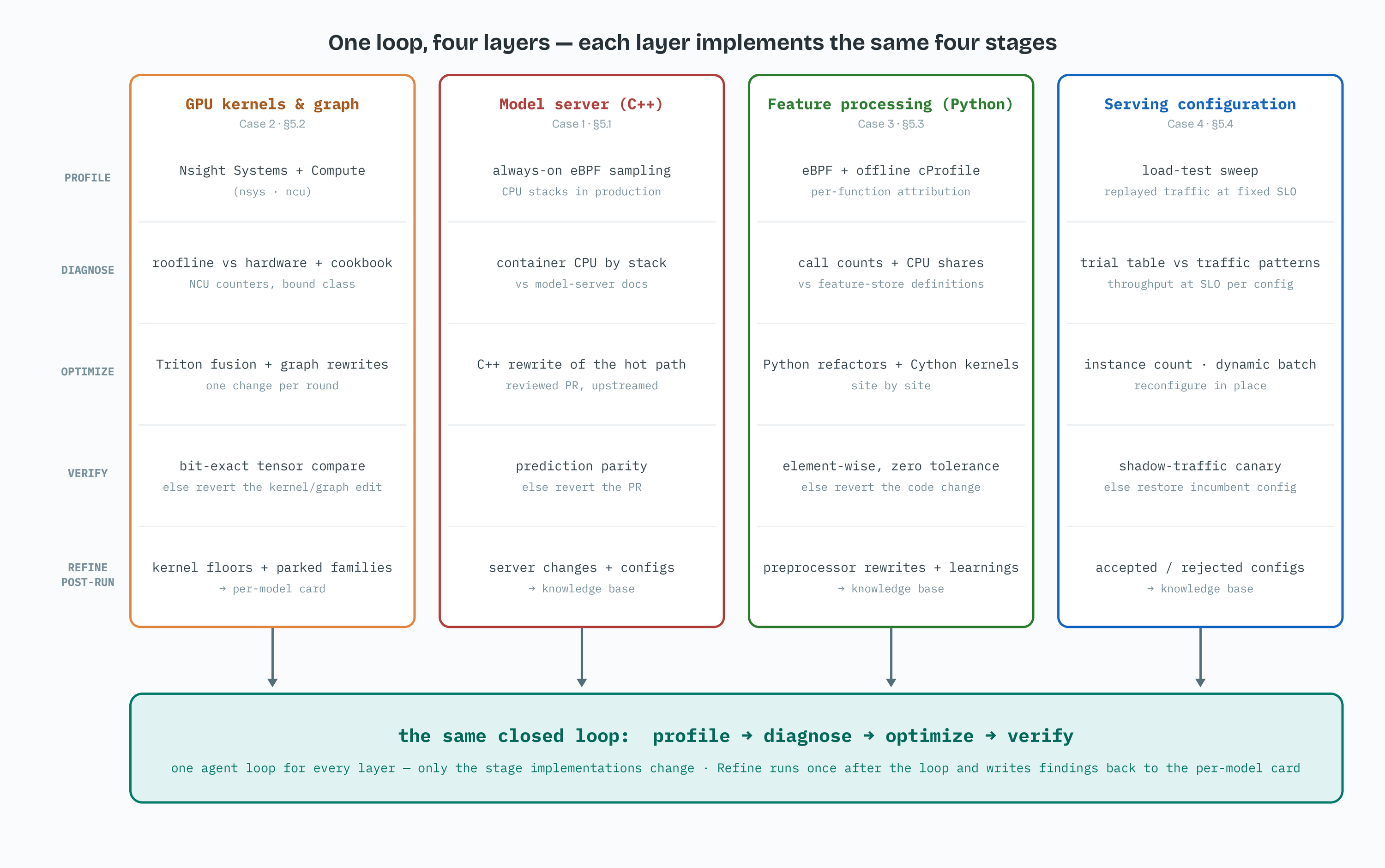}
  \caption{The layer agent abstraction. Every layer implements the same four stages --- Profile, Diagnose, Optimize, Verify --- plus the post-run Refine that writes findings back to the knowledge base; the loop itself is identical across layers. Cell entries name the concrete instance each production layer implements today.}
  \label{fig:layer_abstraction}
\end{figure*}

\subsection{Knowledge Base}
\label{sec:knowledge_base}
FlashVector maintains a knowledge base for the holistic model serving stack, covering all the major components: 

\textbf{Business Logic.} 
Detailed metrics and properties are specialized for Vector's traffic patterns, which directly impact the model's architecture. For example, the number of ad candidates per request, input feature structure, and how gamer-level features are different from candidate-level features.

\textbf{Model Server.} 
Detailed architecture of the model server (NVIDIA Triton server) and runtime parameters (batch size, queue length, etc.). 

\textbf{Feature Store.} 
Detailed data types, lineage, and definitions of the input features to the model server.

\textbf{ML Model.} 
The way in which ML models are implemented in code has a major impact on their inference efficiency. In many situations, machine-learning engineers (MLEs) are unable to express the most hardware-efficient model designs using high-level Python and mainstream ML frameworks such as PyTorch. Therefore, it is crucial to capture and compile detailed model implementation information, such as architecture summary, parameter inventory, and serving configuration, so that an AI agent can identify alternative modules or kernels and select more efficient replacement algorithms.

\textbf{Hardware.} We primarily deploy models on NVIDIA GPUs. Hence, we maintain documentation that outlines standard procedures for profiling and optimizing machine learning programs running on NVIDIA GPUs.

\textbf{Optimization Cookbook.} A curated set of optimization rules and best practices distilled from human experts. This resource is crucial for enabling an AI agent to identify optimization opportunities more efficiently within a limited context window, while ensuring it avoids making erratic or unjustified decisions.

\subsection{Optimization Loop}
\label{sec:optimization_loop}

FlashVector uses a closed-loop workflow comprising four steps: profile, diagnose, optimize, and verify; each iteration samples a single layer and runs that layer's own instance of the four steps, rather than sweeping every layer at once. The refine stage is not a loop step but runs once after the loop terminates, writing what was learned back into the knowledge base (the knowledge-refinement stage of \Cref{fig:figure2}). The workflow is described in \Cref{alg:flashvector}. 

\begin{algorithm}[t]
  \caption{FlashVector optimization loop}
  \label{alg:flashvector}
  \begin{algorithmic}[1]
  \Require Target model serving stack $S$ with layers $\mathcal{L} = \{\ell_1, \dots, \ell_n\}$ (GPU kernels, computation graph, model server, on-demand feature processing), workload definition $W$, evaluation harness $H$, knowledge base
  $K$, noise floor $\epsilon$ estimated from repeated baseline runs
  \Ensure Optimized code/configuration $C^*$ and updated knowledge base $K^*$
  \State $C \gets$ Current source code/runtime configuration for $S$
  \State $K_S \gets \textsc{RetrieveContext}(K, S)$
  \Repeat
    \State $\ell \gets \textsc{SampleLayer}(\mathcal{L})$ \Comment{pick one layer at random this iteration}
    \State $r \gets \textsc{Profile}(\ell, C, W)$ \Comment{profile only $\ell$}
    \State $d \gets \textsc{Diagnose}(r, K_S[\ell])$ \Comment{rank $\ell$'s own bottleneck hypotheses, return the top one}
    \State $\Delta \gets \textsc{Plan}(d)$ \Comment{Optimize (i): propose an edit for $\ell$}
    \State $C' \gets \textsc{Apply}(C, \Delta)$ \Comment{Optimize (ii): apply it}
  \State $(ok, m) \gets \textsc{Verify}(H, C')$ \Comment{correctness and latency/throughput/cost metrics for $\ell$}
  \If{$ok$ \textbf{and} $m$ improves on the incumbent by more than the measurement noise $\epsilon$}
    \State $(ok_{\text{stack}}, m_{\text{stack}}) \gets \textsc{FullStackLoadTest}(H, C')$ \Comment{confirm no end-to-end regression across the full stack}
    \If{$ok_{\text{stack}}$}
      \State $C \gets C'$ \Comment{accept update}
    \Else
      \State discard $C'$ \Comment{local gain regressed the full stack; revert}
    \EndIf
  \Else
    \State discard $C'$ \Comment{revert to the incumbent $C$}
  \EndIf
  \Until{Maximum number of iterations reached \textbf{or} optimization criterion satisfied.}
  \State $C^* \gets C$
  \State $K^* \gets \textsc{RefineContext}(K, S, C^*)$ \Comment{Post-run refine stage}
  \State \Return $(C^*, K^*)$
  \end{algorithmic}
\end{algorithm}

\textbf{Profile}
Rather than focusing on a single GPU kernel optimization problem like most Kernel Design Agents~\cite{KernelAgent2025}, FlashVector collects performance profiles across the model serving stack: GPU kernels, the ML-framework computation graph, the model server, and on-demand feature processing. It runs profiles in environments of increasing complexity, ranging from a local simulated environment (for GPU kernels) to a production-like model-serving load test (for model servers and on-demand feature processing). 

Concretely, this step launches a single process for the layer sampled this iteration and loads the agent skill customized for profiling within that layer's technical domain. The GPU kernel optimization layer agent loads an agent skill specialized in GPU profiling (NVIDIA Nsight Systems and Nsight Compute), runs the benchmark, and collects nsys/ncu reports. The model server optimization agent loads an agent skill specialized in CPU profiling of the inference model server (eBPF), runs the load tests, and collects Python/C++ profiles. Finally, the optimization agent emits a structured report of the profiles. 

\textbf{Diagnose}
The diagnose step reasons about the profile produced for the sampled layer and identifies its bottlenecks. Because each layer's report differs in shape and vocabulary, this step invokes the agent and context customized for that layer: for a GPU kernel, the agent is given the knowledge base focused on GPU and CUDA; for the ML framework (e.g., computation-graph level), it is fed the model definition, the ML framework's source (e.g., PyTorch), and the knowledge base focused on business logic; for the model server, it receives the model server's source repositories and the knowledge base for the model server. Diagnosis ends by ranking the sampled layer's own hypotheses: each candidate is scored by its share of total time multiplied by its expected gain, and one whose best possible gain is smaller than the measurement noise is skipped. The top-ranked candidate is handed to Optimize.

\textbf{Optimize (Plan \& Apply)}
The optimization step generates candidate code modifications based on the bottlenecks identified during diagnosis. Depending on the layer where the bottleneck originates, FlashVector loads domain-specific agents that formulate an optimization plan and translate it into code changes. Although all layer-specific agents share a uniform integration interface, their implementation complexity varies significantly. For example, the GPU kernel agent is relatively straightforward, as it typically modifies a single file. Conversely, the model server agent is considerably more sophisticated, as it operates on larger repositories and manages complex server runtime. To simplify the model server agent implementation, we merged several repositories into a mono repo which is more agent friendly. 

\textbf{Verify} 
Performance improvements are only meaningful if prediction quality is preserved. Certain optimizations sacrifice accuracy in favor of performance, such as approximate algorithms or lossy compression. To protect against this, we verify that such changes remain acceptable with respect to business metrics before proceeding to the next optimization round.

Before the optimization loop, FlashVector records all inference-related artifacts, including model weights, representative input samples, and the corresponding prediction outputs. During verification, FlashVector reloads the recorded weights and inputs into
the optimized stack and compares the resulting predictions against the
recorded baseline. The bar depends on the class of change:

\begin{itemize}
\item \textbf{Deterministic rewrites} (operation order and accumulation
  precision preserved): outputs must be exactly equal, element by element.
\item \textbf{Reordering floating-point rewrites}: end-to-end predictions
  must stay within $10^{-3}$ element-wise absolute deviation, with at most
  $0.1\%$ of elements beyond $10^{-3}$ (one FP16 ULP at unit scale;
  $10^{-4}$ for FP32 serving); non-finite outputs fail outright. The
  per-kernel verifier applies dtype-scaled bounds ($10^{-3}$ FP32,
  $10^{-2}$ FP16) that a change may tighten but never loosen; accepted
  thresholds are written back to the model card.
\item \textbf{Intentional approximations} (outputs differ by design): gated
  on business metrics instead of output equality.
\item \textbf{Configuration changes} (predictions untouched by construction):
  gated by the shadow-traffic canary at a fixed latency SLO
  (\Cref{sec:case4}).
\end{itemize}

Correctness alone does not accept a change: its improvement must also exceed
the noise floor $\epsilon$ of \Cref{alg:flashvector}, estimated from repeated
no-op runs of the unmodified baseline ($\pm 6\%$ end to end); improvements in
the $2$--$6\%$ band are re-measured at up to $8\times$ the sample size,
tightening the floor as $1/\sqrt{k}$ ($2.1\%$ at $8\times$).

Because each iteration profiles and verifies only the sampled layer, a locally verified change could still regress the system once combined with the rest of the stack. Every candidate that clears both the correctness bar and the noise floor is therefore checked once more with a full-stack load test (\textsc{FullStackLoadTest} in \Cref{alg:flashvector}) before being committed, confirming end-to-end throughput and latency hold under production-representative load; a regression there discards the candidate even though it passed locally.

\textbf{Refine} In the refinement stage, the system aggregates the modifications introduced over the course of the optimization run—such as updates to model implementations, model server configurations, and other changes—as well as revisions to the knowledge base described in \Cref{sec:knowledge_base}. FlashVector also records a brief summary documenting these changes along with any performance improvements or regressions observed during the run. 

The refinement stage plays a crucial role in determining the work quality of FlashVector. By applying in-place updates to the knowledge base at the end of each run, FlashVector bounds the amount of memory the agent maintains. Compared with continually adding full optimization memories at each step, this strategy greatly reduces the size of the context window. In practice, it substantially mitigates catastrophic forgetting and hallucinations in FlashVector. 

\textbf{Triggering.} The loop runs without a person invoking it. The model serving stack is optimized whenever a new release enters the fleet, and also when it is still serving traffic but has not been analyzed in long enough that its configuration can no longer be assumed current. This keeps optimization from decaying between manual passes as models retrain, traffic shifts, and infrastructure changes (\Cref{sec:challengs}).
  
\section{Evaluation \& Case Studies}
\label{sec:case_studies}
In this section, we share a few case studies of real world full stack optimizations that FlashVector delivered. \Cref{tab:prod_results} summarizes the resulting speedups at a glance. Not only is it able to find unusual GPU kernel optimizations but can also pinpoint bottlenecks that are not commonly looked at automatically and systematically. 

The baseline measurements and all subsequent profiling of optimized variants were carried out in the same environment to maintain consistency and ensure fair comparison.

\begin{table}[th]
    \centering
    \small
    \begin{tabular*}{\linewidth}{@{\extracolsep{\fill}}lrr@{}}
      \toprule
      Model & Latency Speedup & Throughput Increase \\
      \midrule
      Retrieval model & 1.67$\times$ & 1.10$\times$ \\
      Ranking model 1 & 1.36$\times$ & 1.80$\times$ \\
      Ranking model 2 & 1.98$\times$ & 2.00$\times$ \\
      Ranking model 3 & 1.30$\times$ & 1.33$\times$ \\
      Ranking model 4 & 1.30$\times$ & 1.35$\times$ \\
      Gamer model     & 1.34$\times$ & 1.15$\times$ \\
      \bottomrule
    \end{tabular*}
    \caption{Model Serving Stack optimizations delivered by FlashVector on production models.}
    \label{tab:prod_results}
\end{table}

\subsection{Case 1: Optimize Model Server} \label{sec:case1}
\begin{figure}[t]
  \centering
  \includegraphics[width=\linewidth]{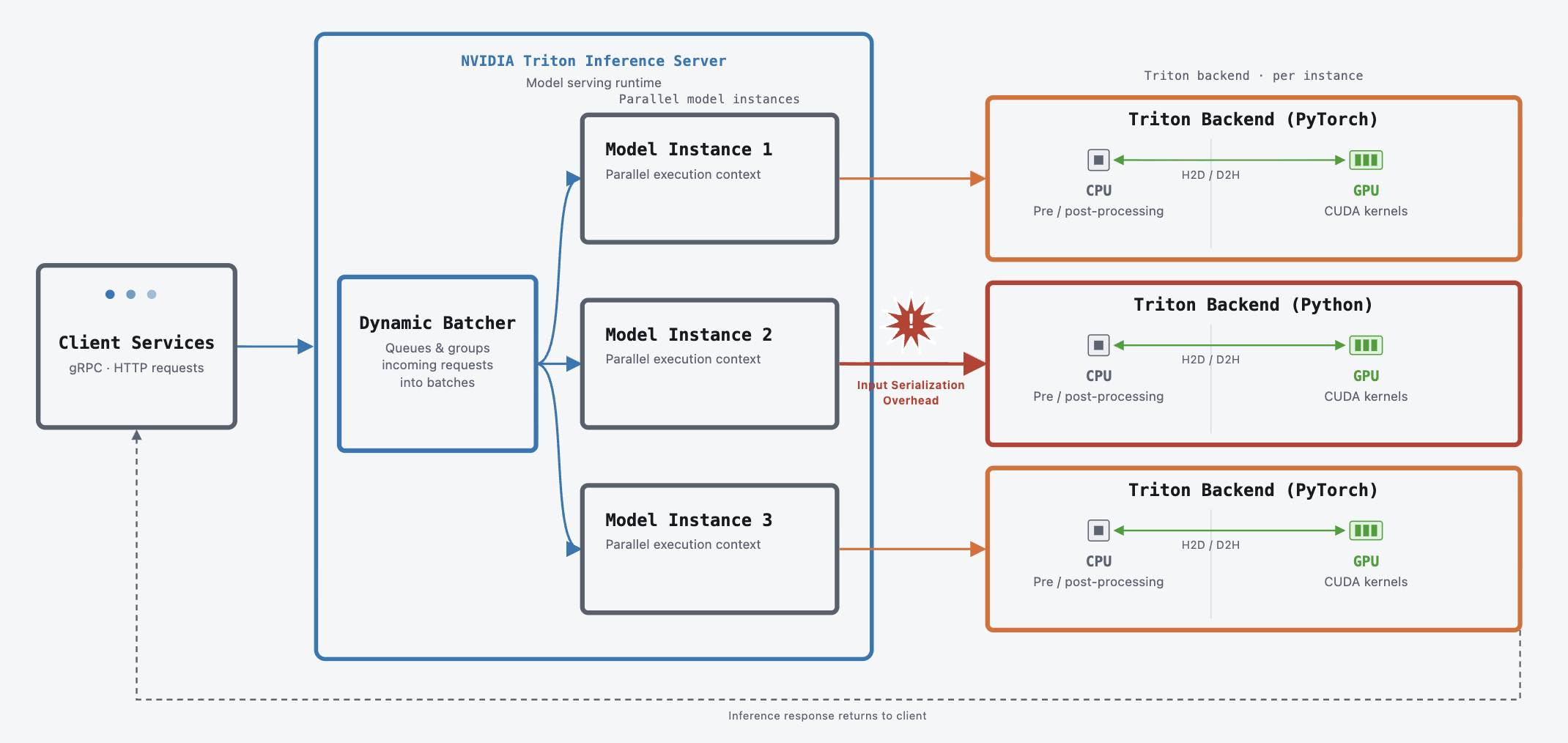}
  \caption{Input Serialization bottleneck found between Triton Server and its Python backend}
  \label{fig:case1}
\end{figure}

For an early stage two-tower retrieval model, as we can see in Figure~\ref{fig:case1}, the FlashVector diagnosis step identified that the bottleneck lies in one step in the NVIDIA Triton Inference Server. The specific step is in converting the server inputs (Triton's gRPC server received over the wire from the clients) into another format that the Triton Python backend (a separate parallel process) expects. However, the conversion logic is implemented in Python's less optimized deserialization module. As the model uses more string inputs, the slow down in converting these string inputs manifested as a major bottleneck. The FlashVector Optimize step hence suggests rewriting the logic in C++. The verification step confirmed the speed up to \textbf{30$\times$} and presented the optimization for review. This case study illustrates that FlashVector can dynamically identify bottlenecks across the stack rather than just narrowly focusing on a specific area, and it can adapt as the workload shifts its bottlenecks. The fix was contributed back \cite{triton_issue_8348} to the NVIDIA Triton Inference server project. 

\FloatBarrier
\subsection{Case 2: Improve Model Inference Efficiency} 
\label{sec:case2}
Another model is a multi-task recommendation model. Dense features pass through a feature-gating layer that adaptively adjusts their contributions, categorical features are represented through shared, learned embedding tables, and target-aware attention modules condense the user’s past install and click sequences in relation to the candidate game. These feature representations are then fed into a stack of feature-interaction layers that capture higher-order relationships among all input signals. The combined representation is passed into a shared bottom MLP, which then splits into several task-specific output heads.

\subsubsection{Baseline setup}

The baseline is AOT-compiled with PyTorch AOTInductor, FP16 quantized, and we use NVIDIA RTX PRO 6000 Blackwell GPU to run all workloads. The model serving payload has batch size of 2000 per request.

\subsubsection{Optimization Highlight}
Through an iterative optimization cycle, FlashVector introduced numerous improvements spanning the kernel and model. Here we highlight several of the most notable optimizations.

\textbf{Fused attention.} 
Standard multi-layer perceptron (MLP) attention layers are bottlenecked by GPU memory bandwidth. Evaluating every candidate–position pair layer-by-layer requires continuously writing and reading hundreds of megabytes of intermediate tensors to GPU memory. FlashVector eliminates this overhead by precomputing candidate-only and position-only feature blocks prior to interaction, avoiding wide per-pair input matrices entirely.

\begin{figure}[t]
  \centering
  \includegraphics[width=\linewidth]{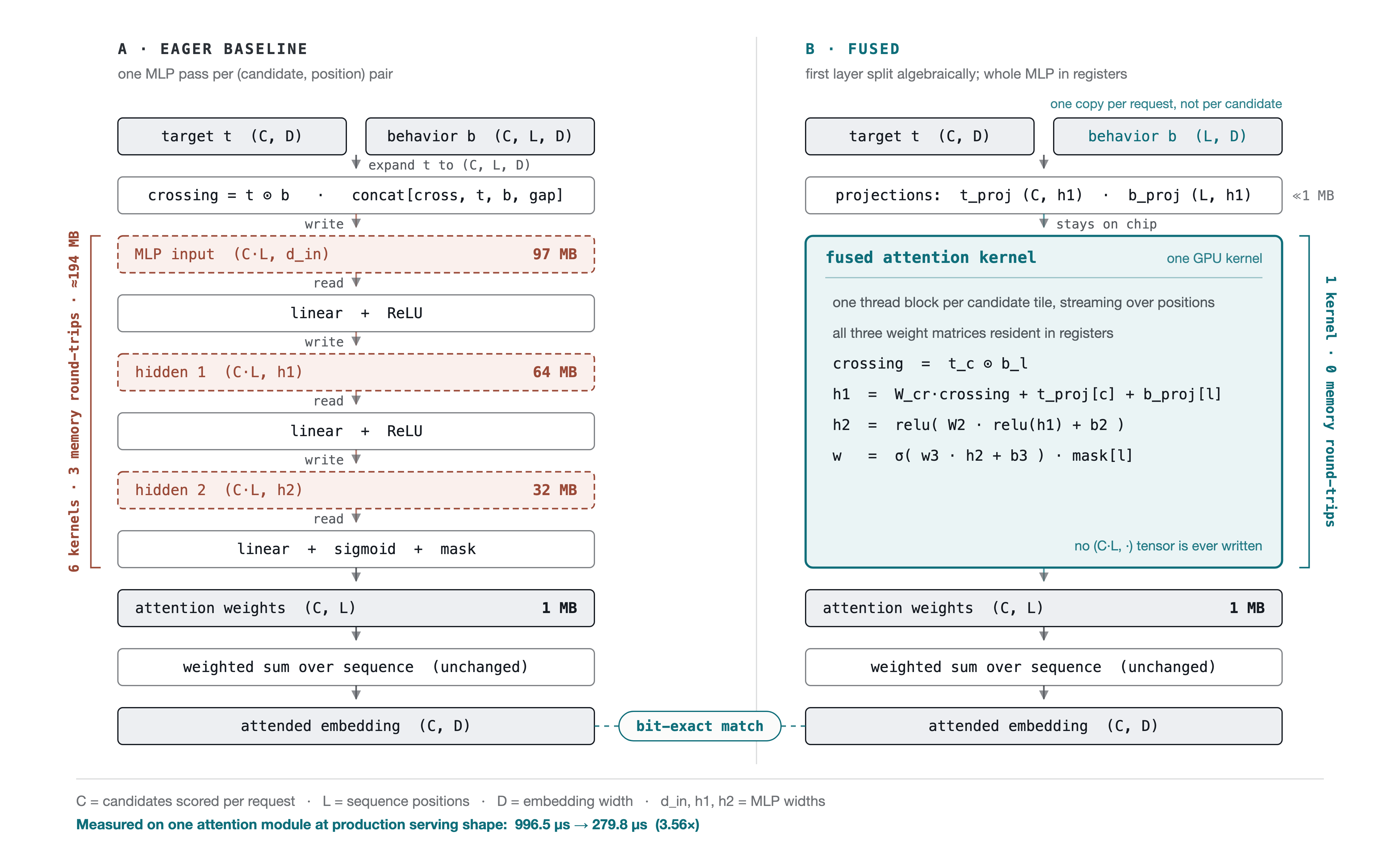}
  \caption{Fused Attention Kernel}
  \label{fig:case3}
\end{figure}

This structure enables consolidating six separate GPU kernels into a single kernel that maintains all intermediate working values within GPU registers (\Cref{fig:case3}). Because the fused kernel preserves the original operation order and accumulation precision, it generates attention weights that are bit-identical to the baseline, which we verify by element-wise comparison. Overall, this transformation reduces the per-module memory traffic to roughly one-third, cutting the attention latency from 996.5\,\textmu s to 279.8\,\textmu s (a $3.56\times$ speedup).

\begin{figure}[t]
  \centering
  \includegraphics[width=\linewidth]{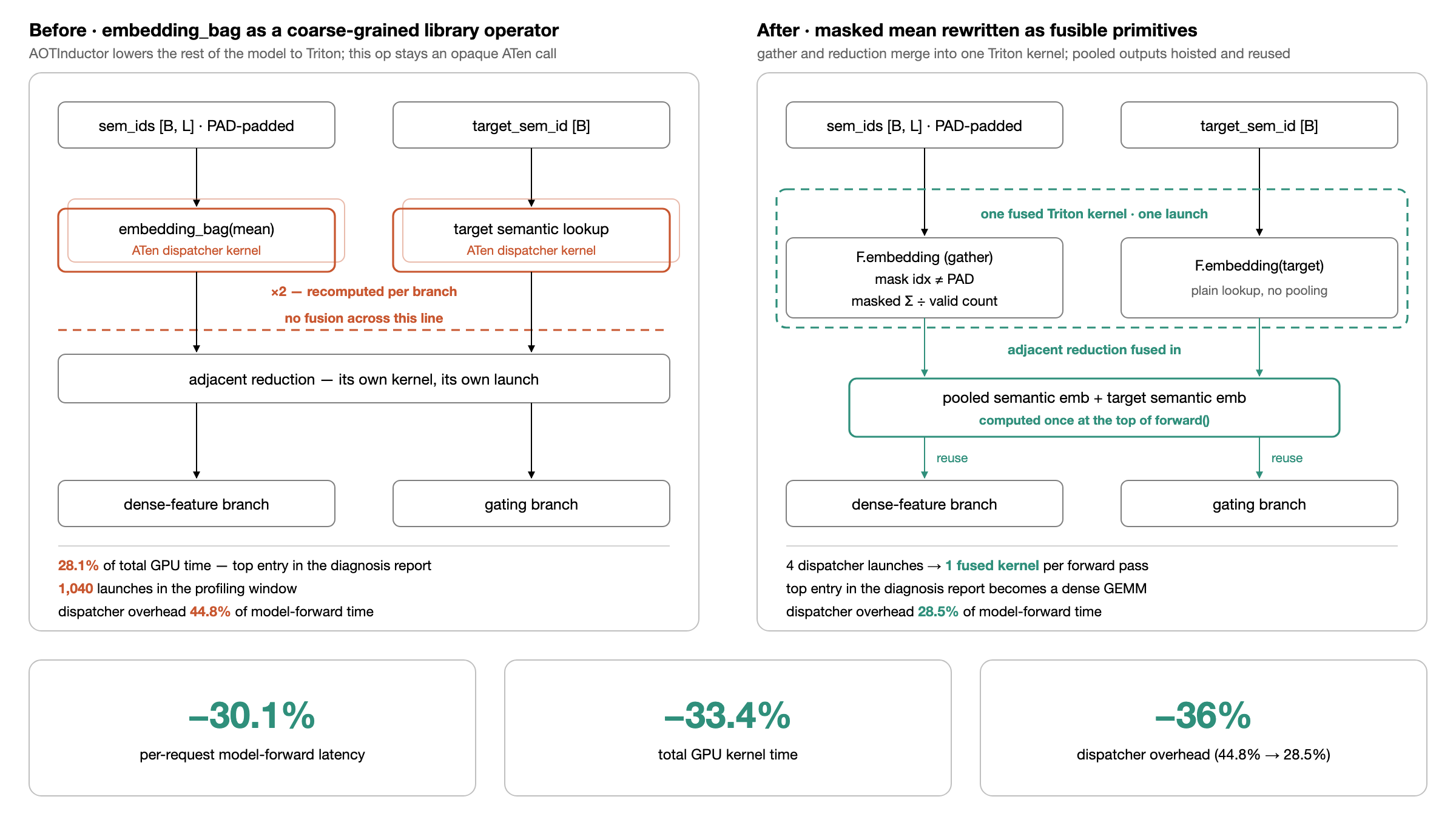}
  \caption{Optimized Embedding Layer}
  \label{fig:case4}
\end{figure}

\textbf{Embedding Layer.} 
FlashVector Diagnose stage finds that the embedding bag kernel consumed 28.1\% of GPU time. Because PyTorch mapped this operator to an opaque ATen dispatcher kernel, AOTInductor was unable to fuse it with adjacent reductions, leading to unnecessary launch overhead and unoptimized subgraphs. FlashVector resolves this by decomposing masked-mean pooling into fine-grained PyTorch primitives (F.embedding, masking, and summation). This exposes the entire lookup and reduction logic to the compiler, which fuses them into a single Triton kernel (\Cref{fig:case4}).

Additionally, FlashVector Optimize Stage eliminates redundant computations by evaluating target and sequence semantic embeddings once at the beginning of the forward pass, reusing the outputs across both dense and gating branches. Combined, these changes reduce the total GPU kernel time by 33.4\%, cut forward-pass latency by 30.1\%, and lower dispatcher overhead from 44.8\% to 28.5\%—shifting the primary execution bottleneck from embedding lookups to matrix multiplication (GEMM) kernels.

\FloatBarrier
\subsection{Case 3: Optimize On-demand Feature Processing} 
\label{sec:case3}

While Case 2 optimizes models deployed on GPUs, on-demand feature processing generates the inputs those models require. For every request, a CPU-only service converts raw request attributes into model input tensors --- decoding and reshaping them, enriching them with additional metadata. This \emph{preprocessor} is deployed as its own Nvidia Triton model through Python backend.

FlashVector loads optimization agent for model on-demand feature processing, of which the Diagnosis step reads eBPF profile directly from production that covers all the processes (Python runtime, Triton's threads), then runs custom scripts to parse it into a structured report and identify the layer's bottlenecks.

\subsubsection{Baseline setup}

The system in question is a preprocessor that handles requests sequentially, where each request includes one context row and as many as 2000 candidate rows. In this experiment, a single pod is running 12 Python backend instances, each in its own process, and they handle requests in parallel across the pod’s 14 vCPUs.

\subsubsection{Optimization highlight}

Diagnosis partitioned the cost along the request's own path --- receiving the inputs, computing the features, and assembling the output tensors, and each stage wasted CPU in its own way.

\textbf{Receiving the inputs.} Before any transformer runs, the Python backend materializes every request tensor as Python objects: the stock accessor copies the server's buffer, and a string tensor becomes one \texttt{bytes} object per element plus a decode loop. A production request carries 875 such inputs, most of them numeric and read only to be reassembled into tensors. FlashVector Optimize stage proposes several optimizations. \emph{Native view} removes the copy: a numeric tensor is exposed as a read-only view onto the server's own buffer, and a string tensor is walked once in C++ rather than one object per element plus a decode loop. \emph{Native mapping} removes the read entirely where an input's only consumer is a vocabulary lookup --- C++ probes the shared-memory vocabulary and the text never reaches Python. Each input's route is resolved once at initialization from the feature graph, and any route that cannot be proven safe falls back to decoding. \Cref{sec:case1} removed the server's own serialization of these inputs; this removes the backend's materialization of them.

\textbf{Computing the features.} Where per-element work is too irregular for vectorized NumPy/pandas operations — ragged loops, per-row dictionaries, string handling — FlashVector's Optimize stage replaces it with hand-written Cython kernels (see \Cref{tab:cython_results} for kernels it produced and speedups). 

\textbf{Assembling the tensors.} Turning per-feature values into the model's input tensors runs in two passes and accounts for roughly 20\% of serving CPU. Both passes waste it the same way — allocating a small array per feature and then concatenating — and because the second pass begins by copying the first pass's layout, this doubles across several hundred features. FlashVector's Optimize stage instead allocates each destination once and lets every feature write directly into its own slice, eliminating both concatenations and the redundant copy.

\begin{table}[th]    
    \centering
    \small
    \begin{tabular*}{\linewidth}{@{\extracolsep{\fill}}lrr@{}}
      \toprule
      Python Functions Replaced && Kernal speedup \\
      \midrule
      pd.to\_datetime, once per request   && 350$\times$ \\
      one numpy assignment per feature   && 25.1$\times$ \\
      bytes $\rightarrow$ unicode per input   && 12.4$\times$ \\
      6 universal functions dispatches $\times$ 180 sites   && $\sim$8$\times$ \\
      per-row \texttt{dict} for one membership query   && 4.4--5.8$\times$ \\
      \bottomrule
    \end{tabular*}
    \caption{Native Cython kernels and the operations they replace, with speedup at
the call site}
    \label{tab:cython_results}
\end{table}

\subsubsection{Results}

The three optimizations were implemented and evaluated in different iterations --- the Cython kernels and the assembly rewrite together, then zero-copy request I/O. Each iteration delivered a double-digit gain, and because each is measured on top of the one before it, those gains compound. \Cref{tab:case3-ladder} tracks the throughput one preprocessor pod sustains at a fixed latency SLO: 940 requests per second at baseline, roughly 1.5k with all three in place --- a \textbf{1.6$\times$} improvement. The sequence is the argument for running the loop repeatedly: each pass re-profiles a service the previous pass has already changed, and surfaces whichever bottleneck now dominates.

\begin{table}[t]
\centering
\small
\caption{Throughput per preprocessor pod as each round of
optimizations was tested in production environment. Each row is
measured on top of the one above it.}
\label{tab:case3-ladder}
\begin{tabular*}{\linewidth}{@{\extracolsep{\fill}}lrrr@{}}
\toprule
Stage & RPS / pod & vs.\ prev. & vs.\ base \\
\midrule
Baseline                              &        940 & ---   & ---         \\
\quad + native kernels, assembly      & $\sim$1.2k & +28\% & $1.3\times$ \\
\quad + zero-copy request I/O         & $\sim$1.5k & +25\% & $1.6\times$ \\
\bottomrule
\end{tabular*}
\end{table}

\FloatBarrier
\subsection{Case 4: Tune Serving Stack Parameters}
\label{sec:case4}
The first three cases show FlashVector discovering better code within a layer. This case shows a complementary capability: FlashVector also adaptively retunes runtime configuration — dynamic batching settings, the request queue policy, the number of model instances resident on a GPU, and more — whenever the surrounding stack changes, including after its own code optimizations land. Across the configurations measured for a single, already code-optimized model, sustained throughput at the same latency SLO varied by more than $2\times$, showing how much is left on the table when configuration isn't retuned alongside code.

\subsubsection{Baseline setup}
Each model server pod owns one GPU partition and the replica count is set by an autoscaler. The sweep is written to search an arbitrary set of levers; at present it varies two of them, instance count and dynamic batch size, with the queue policy derived from the SLO rather than searched. A request here is already a ragged batch of roughly a thousand ad candidates packed into one tensor, so a dynamic batch of 6 is a GPU batch of several thousand candidate rows. The hand-tuned baseline left the batch size at the framework default and scaled on a utilization target.

\subsubsection{The loop}
This layer's Profile step runs each candidate configuration under held load and records its sustained throughput at a fixed latency SLO. Diagnose ranks the trials and picks the best-performing configuration, or declines when nothing beats the incumbent by more than the measurement noise. Optimize applies that configuration by opening it as a pull request. Verify tests the change in production through a shadow-traffic canary, adopting it if the canary holds and discarding it if not, with either outcome recorded in the knowledge base. The loop repeats on the next model release. Fidelity comes from replaying real production requests rather than synthetic ones; for a two-stage model the paired preprocessor's outputs are captured first and replayed as the model's inputs, so the tensors are the ones production produces. The job runs isolated on a node of the production machine type.

\subsubsection{Results}
\Cref{tab:case4-results} lists four production models tuned this way. Each run was launched, monitored, interpreted, and delivered as a reviewed pull request by the agent.

\begin{table}[t]
\centering
\small
\begin{tabular*}{\linewidth}{@{\extracolsep{\fill}}lrrrrr@{}}
\toprule
& & \multicolumn{2}{c}{inst.\ / batch} & \multicolumn{2}{c}{chosen trial} \\
\cmidrule(lr){3-4}\cmidrule(lr){5-6}
Model & SLO & base & tuned & RPS & $p99$ \\
\midrule
Model A & 250 & 7 / 64  & 6 / 6 & 1321 & 192 \\
Model B & 250 & 10 / 64 & 7 / 3 &  736 & 271 \\
Model C & 300 & 8 / 2   & 4 / 2 &  238 & 251 \\
Model D & 400 & 12 / 3  & 3 / 1 &  139 & 285 \\
\bottomrule
\end{tabular*}
\caption{Configurations selected by the agent from the sweep results. SLO and $p99$ in ms; RPS is the throughput the chosen
trial sustained at the SLO.}
\label{tab:case4-results}
\end{table}

\section{Discussion and Conclusion}
\label{sec:discussion}

We presented FlashVector, an agentic system that extends the recent success of LLM-driven GPU kernel optimization to the full hierarchical model-serving stack. Rather than treating each layer's optimization as a bespoke, human-expert-driven effort, FlashVector generalizes the profile--diagnose--optimize--verify loop through three mechanisms: a \emph{layer agent abstraction} that lets a kernel, an ML framework, a model server, or a feature-processing service plug into the same closed loop by implementing the Profile, Diagnose, Optimize, and Verify steps of \Cref{fig:layer_abstraction} with its own tooling; \emph{optimize locally, verify globally}, which proposes candidates independently within each layer but accepts one only when it is shown to improve the system as a whole—measured against replayed production traffic—and keeps it only when the gain clears the measurement noise; and \emph{an always-on optimization loop}, which starts each run from the model's recorded history of accepted and rejected changes, writes the run's outcome back, and re-triggers itself automatically—so that discovered optimizations survive model retraining, traffic shifts, and hardware upgrades.

Deployed on Unity's Vector advertising platform, FlashVector delivered up to \textbf{2$\times$} throughput increase and up to \textbf{1.98$\times$} latency speedup on the model server, and up to \textbf{1.6$\times$} throughput increase on the feature store, with optimizations spanning GPU kernel fusion and embedding-layer restructuring (\Cref{sec:case_studies}), model-server input serialization, CPU-bound on-demand feature processing, and online serving-parameter search. That this range of fixes was discovered by the same closed loop, without a new system being purpose-built for each layer, is the paper's main lesson: cross-layer serving inefficiency is not merely a kernel problem in disguise, and the agentic paradigm that has proven effective for kernels generalizes to the heterogeneous, multi-language systems that surround them. The continuous configuration-tuning loop in \Cref{sec:case4} further shows that this generalization extends beyond individual code fixes to a loop that keeps running on its own, continuously, keeping pace with a fleet whose models, traffic, and infrastructure never stop changing.

More broadly, modern system architectures show an increasingly distinct divide between reference and production languages. Although Python and PyTorch are preferred for algorithmic prototyping owing to their developer velocity~\cite{yang2026pytorch}, production systems necessitate superior performance. FlashVector's case studies are, in effect, instances of this divide being bridged automatically: parallel agent workflows translate reference implementations into optimized, lower-level code in CUDA and C/C++, or into equivalent low-level Python and Cython where a full language change is unnecessary, without requiring the engineering team to maintain two parallel implementations by hand. In this view, FlashVector is not only a serving-optimization tool but an early instance of a broader pattern: agentic systems as the connective layer between fast-moving reference code and the performance-critical systems it must eventually become.

\section{Acknowledgement}
We are grateful to our team members for their invaluable contributions to the development of this work: Zhusong Mei, Philippe Reddy, Oscar Elfving, David Bellemare, Sam Hallam, Xianghao Chen, Ryan Ngo, Sonja Li.

\bibliographystyle{mlsys2025}
\bibliography{references}

\end{document}